\documentclass[letterpaper, 10 pt, conference]{ieeeconf}
\IEEEoverridecommandlockouts
\usepackage{cite}
\usepackage{amsmath,amssymb,amsfonts}
\usepackage{algorithmic}
\usepackage{graphicx}
\usepackage{textcomp}
\usepackage{color, soul}
\usepackage{xcolor}
\def\BibTeX{{\rm B\kern-.05em{\sc i\kern-.025em b}\kern-.08em
    T\kern-.1667em\lower.7ex\hbox{E}\kern-.125emX}}

\usepackage{makecell}
\usepackage{graphicx}
\usepackage{textcomp}
\usepackage{array} 
\usepackage{longtable}
\usepackage{booktabs}
\usepackage{float}
\usepackage{dblfloatfix}

\title{\LARGE \bf
Sparse Color-Code Net: Real-Time RGB-Based 6D Object Pose Estimation on Edge Devices
}

\author{Xingjian Yang$^{1}$, Zhitao Yu$^{1}$, and Ashis G. Banerjee$^{2}$%
\thanks{$^{1}$X. Yang and Z. Yu are with the Department of Mechanical Engineering, University of Washington, Seattle, WA 98195, USA.
{\tt\small yxj1995,zhitaoyu@uw.edu}}%
\thanks{$^{2}$A. G. Banerjee is with the Department of Industrial \& Systems Engineering and the Department of Mechanical Engineering, University of Washington, Seattle, WA 98195, USA.
{\tt\small ashisb@uw.edu}}
}

\begin{document}

\maketitle
\thispagestyle{empty}
\pagestyle{empty}

\begin{abstract}
As robotics and augmented reality applications increasingly rely on precise and efficient 6D object pose estimation, 
%the pursuit of 
real-time performance on edge devices is required
%driven by the need 
for more interactive and responsive systems. Our proposed Sparse Color-Code Net (SCCN) embodies a clear and concise pipeline design to effectively address this requirement. %introducing an effective strategy for 6D object pose estimation optimized for real-time operation on edge devices. 
SCCN performs pixel-level predictions on the target object in the RGB image, utilizing the sparsity of essential object geometry features to speed up the Perspective-n-Point (PnP) computation process. Additionally, it introduces a novel pixel-level geometry-based object symmetry representation that seamlessly integrates with the initial pose predictions, effectively addressing symmetric object ambiguities. SCCN notably achieves an estimation rate of 19 frames per second (FPS) and 6 FPS on the benchmark LINEMOD dataset and the Occlusion LINEMOD dataset, respectively, 
%when deployed 
for an NVIDIA Jetson AGX Xavier, while consistently maintaining high 
%pose 
estimation accuracy at these rates. 
%The implementation of SCCN is available at https://github.com/xxxxxx
\end{abstract}

\section{Introduction}
6D object pose estimation, a crucial task in computer vision, has advanced significantly due to deep learning and increasing demand in various applications \cite{brachmann2014learning}. The goal is to determine the 3D orientation and position of objects from 2D images, facing challenges such as occlusions, cluttered scenes, and lighting variations \cite{lepetit2005monocular,lepetit2009ep,kehl2017ssd}. Traditional methods based on geometric features and templates struggle with texture-less objects and heavily occluded or dynamically lit scenes \cite{rothganger20063d,hinterstoisser2013model}. Deep learning approaches, including direct pose regression \cite{xiang2017PoseCNN,manhardt2018deep}, keypoint detection \cite{peng2019pvnet}, and hybrid variants, have improved accuracy and efficiency, especially when addressing the synthetic-to-real domain gap \cite{chen2020naive,zhang2020weakly}. However, the high dimensionality of the pose space and the need for large annotated datasets remain challenging. Promising solutions include using synthetic data with minimal real annotations \cite{zhang2020weakly,chen2020naive}, keypoint-based methods with PnP algorithms \cite{lepetit2009ep,fischler1981random}, segmentation-driven techniques for occlusion handling \cite{park2019Pix2Pose,prisacariu2012pwp3d,hu2019segmentation}, and integrating RGB-D data \cite{kehl2017ssd}.

    \begin{figure}[thpb]
      \centering
      \includegraphics[scale=0.15]{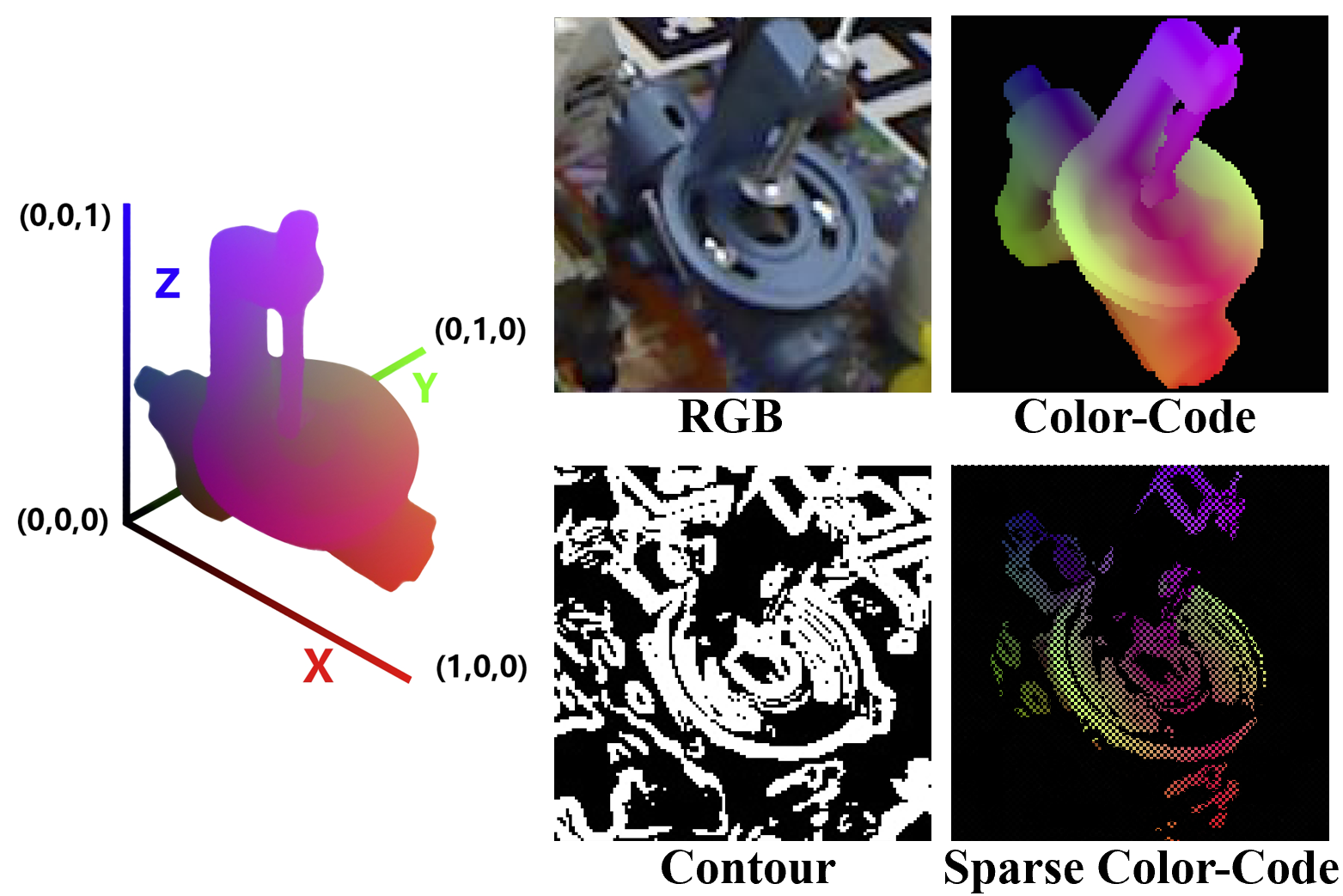}
      \caption{Left: Visualization of the color-code of an object. The object is normalized to a 1$\times$1$\times$1 cube, with its longest dimension spanning the full range. The X, Y, and Z axes map to the R, G, and B color channels, respectively, giving each surface point a unique color based on its position. Right: Illustration of the input RGB image, image contour captured by the Sobel filter, the color-code estimation of the target object, and the sparse color-code representation, which is the final output of the pipeline.}
      \label{brief_illustration}
   \end{figure}

Recent trends focus on end-to-end trainable frameworks that efficiently handle multiple objects and instances \cite{thalhammer2023cope,amini2022yolopose}, predict the 6D poses directly from the RGB inputs, and operate in real-time. Novel data augmentation and network architectures are designed to improve feature utilization, handle occlusions \cite{thalhammer2023cope,castro2023crt}, and minimize computational load. Self-supervised learning from synthetic data \cite{wang2020self6d} and weakly supervised approaches using 2D annotations \cite{zhang2020weakly} aim to overcome data scarcity and labeling costs. The diverse applications of pose estimation often require processing on mobile computing platforms due to limited access to low-latency, high-bandwidth networks. Additionally, many use cases demand real-time performance, which can be challenging given the substantial computational requirements of the current pose estimation models.

    \begin{figure*}[!b]
      \centering
      \includegraphics[scale=0.165]{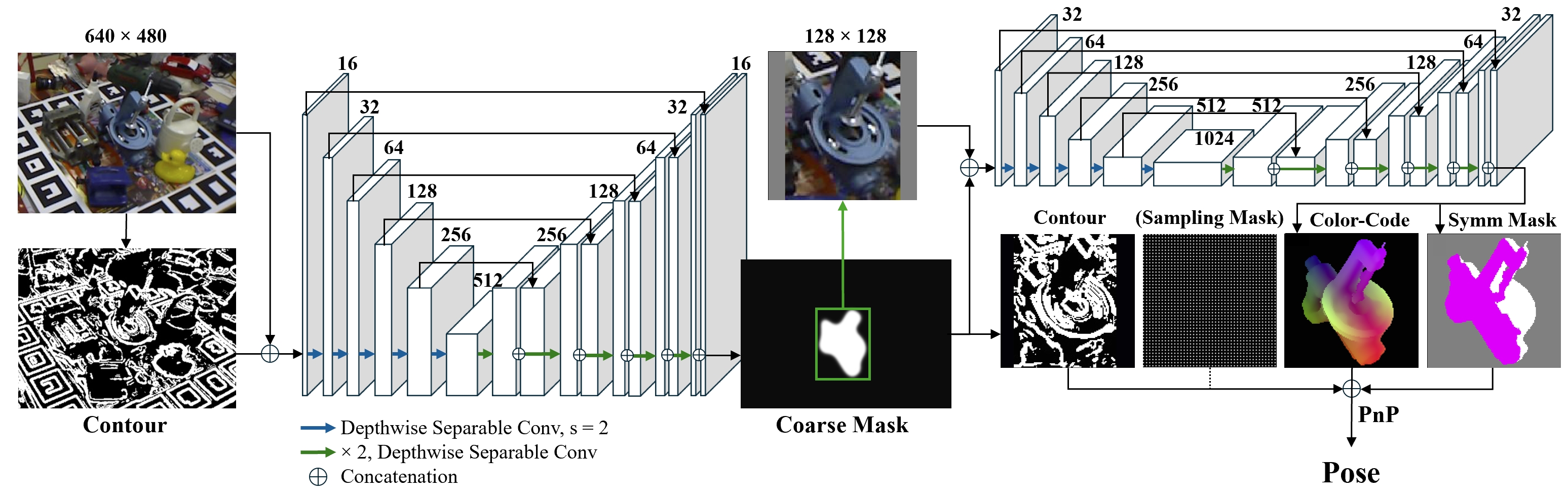}
      \caption{An overview of the Sparse Color-Code Net pipeline. It takes an input RGB image and applies Sobel filters to extract contours. The contours and RGB image are fed into a UNet to generate a coarse object mask. The mask determines a bounding box, which is used to crop, pad, and resize the RGB image, contour, and mask. This combination is the input to another UNet that estimates the color-code and symmetry mask. Finally, with the contour and sampling mask (optional), PnP estimation is used to determine the object's pose.}
      \label{pipeline}
   \end{figure*}

We propose Sparse Color-Code Net (SCCN), a three-stage pipeline for real-time 6D pose estimation of single or multiple objects. The color-code representation (see Fig. \ref{brief_illustration}) provides an intuitive and straightforward approach for neural networks to 
%inherently 
learn and memorize the correspondences between object points and their associated colors. Building upon prior work \cite{park2019Pix2Pose}, SCCN (see Fig. \ref{pipeline}) introduces key improvements for enhanced accuracy and efficiency. The first stage employs Sobel filters to extract sparse contours representing important surface details. The input image and contours are passed to a UNet \cite{ronneberger2015u}, which segments target objects and locates their bounding boxes. In the second stage, cropped object patches are processed by another UNet for pixel-level color-code regression, establishing 2D-3D correspondences and predicting a novel symmetry mask to resolve ambiguities. Finally, color-code pixels are selected based on the contours and transformed into a 3D point cloud. The PnP algorithm generates the final 6D pose estimate. By focusing on important regions and leveraging sparsity, SCCN achieves high speed without significantly compromising accuracy.

We, therefore, summarize our contributions 
%in this work 
as follows:
\begin{itemize}
  \item Our sparse color-code predictions and pipeline optimizations enable high-speed 6D pose estimation without significantly compromising accuracy.
  \item We introduce a novel pixel-level representation to resolve ambiguities in symmetric objects, enabling robust pose prediction regardless of object symmetry.
    \item Our framework outperforms many end-to-end approaches for both single and multiple objects on an NVIDIA Jetson Xavier platform in terms of efficiency. 
    %with decent accuracy.
\end{itemize}

\section{RELATED WORK}

\textbf{Classical Approaches.} Traditional methods rely on hand-crafted features like SIFT to match the images with 3D models \cite{lowe2004distinctive}. These features are designed to be invariant to scale, rotation, illumination, and viewpoints. RANSAC handles the outliers \cite{phillips2016seeing}, and PnP recovers the pose \cite{prisacariu2012pwp3d}. However, these methods face difficulties with textureless and occluded objects, since local feature matching requires sufficient image textures \cite{rothganger20063d}. RGB-D methods incorporate depth data, using template matching \cite{hinterstoisser2011multimodal}, Hough voting \cite{hinterstoisser2016going}, and algorithms leveraging both color and depth gradients.

\textbf{CNN-based RGB Pose Estimation.} Recently, convolutional neural networks (CNNs) have dominated 6D object pose estimation, using either direct pose regression or 2D-3D correspondences for the PnP computation. Regression maps images to pose parameters, but face challenges with loss functions over 3D rotations and symmetric object ambiguities. Correspondence methods predict the 2D projections of the coordinates or keypoints to match the 3D models \cite{tekin2018real,peng2019pvnet,zakharov2019dpod}, avoiding regression difficulties by decomposing the problem. They leverage mature 2D detectors like YOLO and Mask R-CNN and
%Keypoint correspondence methods 
demonstrate high accuracy \cite{li2019cdpn, hu2019segmentation, park2019Pix2Pose}. Voting schemes improve occlusion robustness by allowing the visible portions to contribute to keypoint localization \cite{peng2019pvnet}. However, global representations still encounter difficulties with truncated and occluded objects, as compared to part-based algorithms \cite{hosseini2019ipose}. 

\textbf{CNN-based RGB-D Pose Estimation.} RGB-D CNN methods combine color and depth for accuracy and handling textureless objects. Earlier works focus on refinement, predicting an initial RGB pose, and optimizing it by aligning the depth data to the renderings \cite{xiang2017PoseCNN,hosseini2019ipose,kehl2017ssd}. Newer approaches explore early and late fusion to better utilize the correlations between the modalities \cite{brachmann2014learning,krull2015learning,michel2017global,wang2019densefusion}. Late fusion concatenates depth maps as additional channels into the RGB pipelines \cite{brachmann2014learning,krull2015learning,michel2017global}, while early fusion uses heterogeneous architectures to process and integrate RGB and depth streams before final pose estimation \cite{wang2019densefusion}. While demonstrating state-of-the-art performance on benchmark datasets, RGB-D methods often have stricter computational requirements, and rely on sensors with high power consumption and sensitivity to background lighting.

\textbf{Attention mechanisms.} Recently, attention and transformer models are emerging as alternatives to CNNs for pose estimation \cite{iftekhar2022look,he2022fs6d,castro2023crt}. Transformers capture longer-range spatial relationships in images \cite{iftekhar2022look}. Attention focuses computation on important regions while retaining that global context. This improves efficiency and robustness. Initial works demonstrate promising refinement and category-level pose estimation capabilities with transformers \cite{he2022fs6d}. However, transformers remain computationally expensive, requiring future optimizations for real-time inference.

\section{METHODS}

Given a single RGB image, 6D object pose estimation aims to determine the full rigid transformation between the object and camera coordinates, consisting of a rotation matrix and a translation vector. Effectively tackling this problem requires exploiting both textural image cues as well as geometric properties.

Our pipeline first applies segmentation to localize objects and extract regional patches encapsulating them (Fig. \ref{pipeline}). A UNet-based architecture then segments these patches to obtain the masks and bounding boxes delineating the target objects. The pipeline subsequently processes cropped object regions in batches, resizing and padding them to standardized dimensions. This batch representation feeds into a secondary network performing dual-task dense pixel-wise prediction: (1) establishing 2D-3D correspondences via color-code regression, and (2) resolving symmetrical ambiguities by predicting a per-pixel symmetry encoding. By combining segmentation and sparse color-code inferences concentrated in highly-informative areas, we obtain refined 2D-3D matches. Transforming these pixel locations into a 3D point cloud representation using known object parameters and symmetry properties then allows training a compact rotation and translation regression network. This network applies convolutional feature extraction followed by fully-connected prediction layers to directly estimate the full 6D pose in a single forward pass. 
%The proposed pipeline aims to balance accuracy and efficiency for real-time multi-object 6D pose recovery from monocular RGB imagery by judiciously leveraging geometrical and perceptual knowledge along with tailored deep networks. 
Fig. \ref{pipeline} depicts the corresponding workflow. We now describe each module individually in the remainder of this section.

\subsection{Object Segmentation} 
Both the segmentation and color-code estimation modules employ UNet-style architectures. Compared to the original UNet, these modified versions retain the downsample and upsample paths with skip connections from the encoder to decoder at identical spatial resolutions. However, the 2D convolution layers in the vanilla UNet are replaced with depthwise separable convolutions to improve computational efficiency. Moreover, the convolutions in the encoder and decoder are designed symmetrically - the encoder utilizes pointwise followed by group convolutions for downscaling (group number equals to the input channel number), while the decoder reverses this order for upsampling. The filter size of group convolution in the encoder is fixed to \(5\times5\) with a stride of 2 to downsample and aggregate the broader spatial context. For the decoder, upsampling is performed via bilinear interpolation, followed by \(3\times3\) group convolutions to refine local feature representations and preserve neighborhood connections. This architecture accelerates training and inference without compromising model capacity much.

\begin{figure}[t]
      \centering
      \includegraphics[scale=0.15]{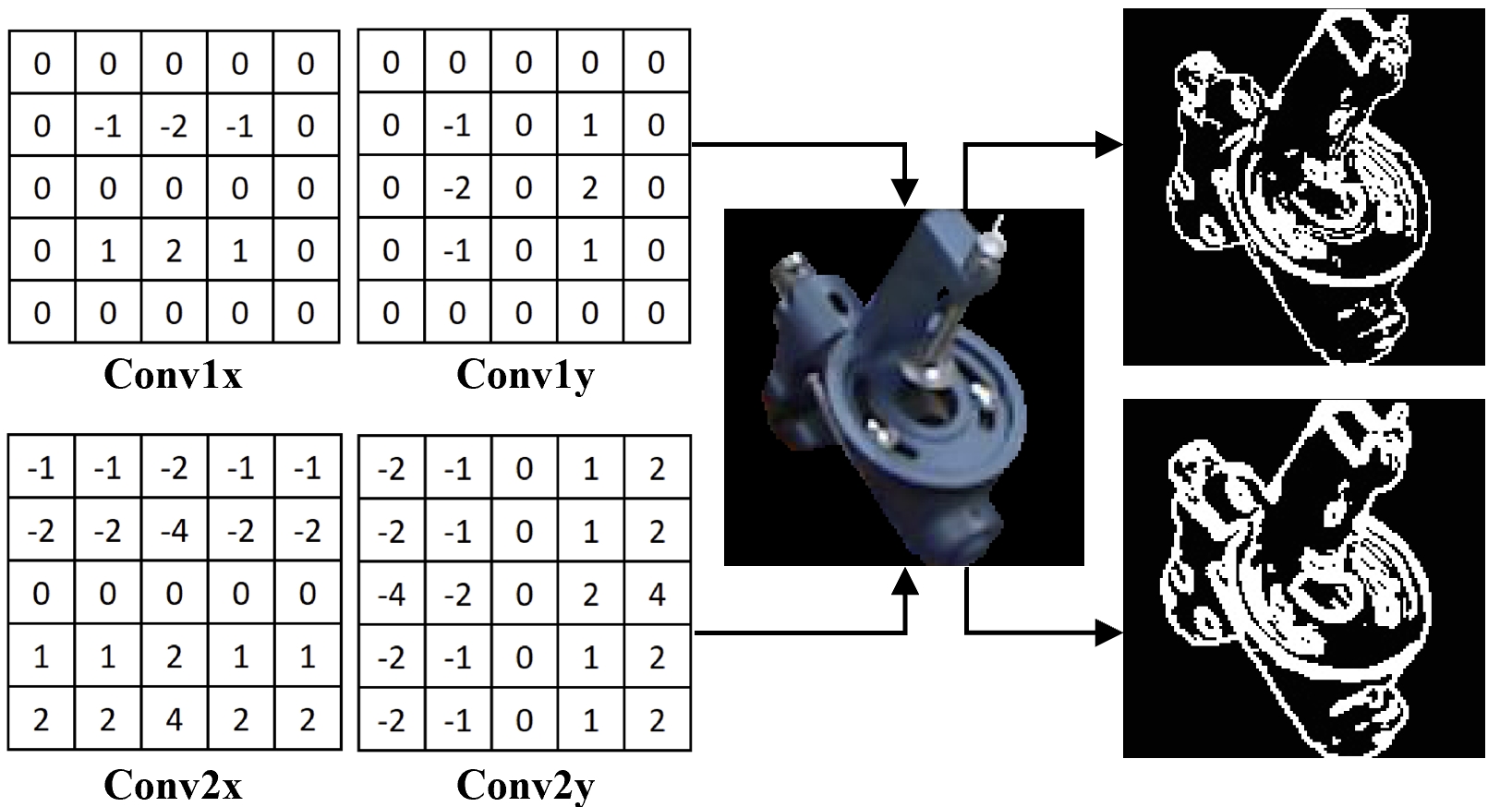}
      \caption{The Sobel filters used to extract the contour. Two sets of Sobel filter kernels: \(3\times3\) (pad to \(5\times5\)) captures finer details and \(5\times5\) captures more general and broader boundary information.}
      \label{sobelfilter}
    \end{figure}

The input scene image is first preprocessed to extract the object contours. This is achieved by filtering with a combination of \(3\times3\) and \(5\times5\) Sobel filters in both horizontal and vertical orientations (Fig. \ref{sobelfilter}), capturing both fine and coarse scale edge details. These contours correspond to the visible surface discontinuities on the object, such as changes in material, texture, edges, and uneven topology.

The segmentation loss function comprises both Cross Entropy and Tversky \cite{salehi2017tversky} losses. Cross Entropy is well-suited for addressing class imbalance and initially provides dominant gradients during early training when the module is still learning coarse region estimation. As convergence improves and the network begins resolving finer segmentation boundaries, the Tversky component escalates in influence. This hybrid loss function is, therefore, defined as:
\begin{equation}
\label{eq:loss_seg}
Loss_{seg} = \lambda_{Tversky} \cdot Loss_{Tversky} + \lambda_{CE} \cdot Loss_{CE}
\end{equation}
where $Loss_{Tversky}$ is the Tversky loss, $Loss_{CE}$ is the Cross Entropy loss. $\lambda_{Tversky}$ and $\lambda_{CE}$ are the weight factors for the two loss terms, and are both set to 1 in our case.

\subsection{Optimal Mask Generation} 
\begin{figure}[!b]
      \centering
      \includegraphics[scale=0.17]{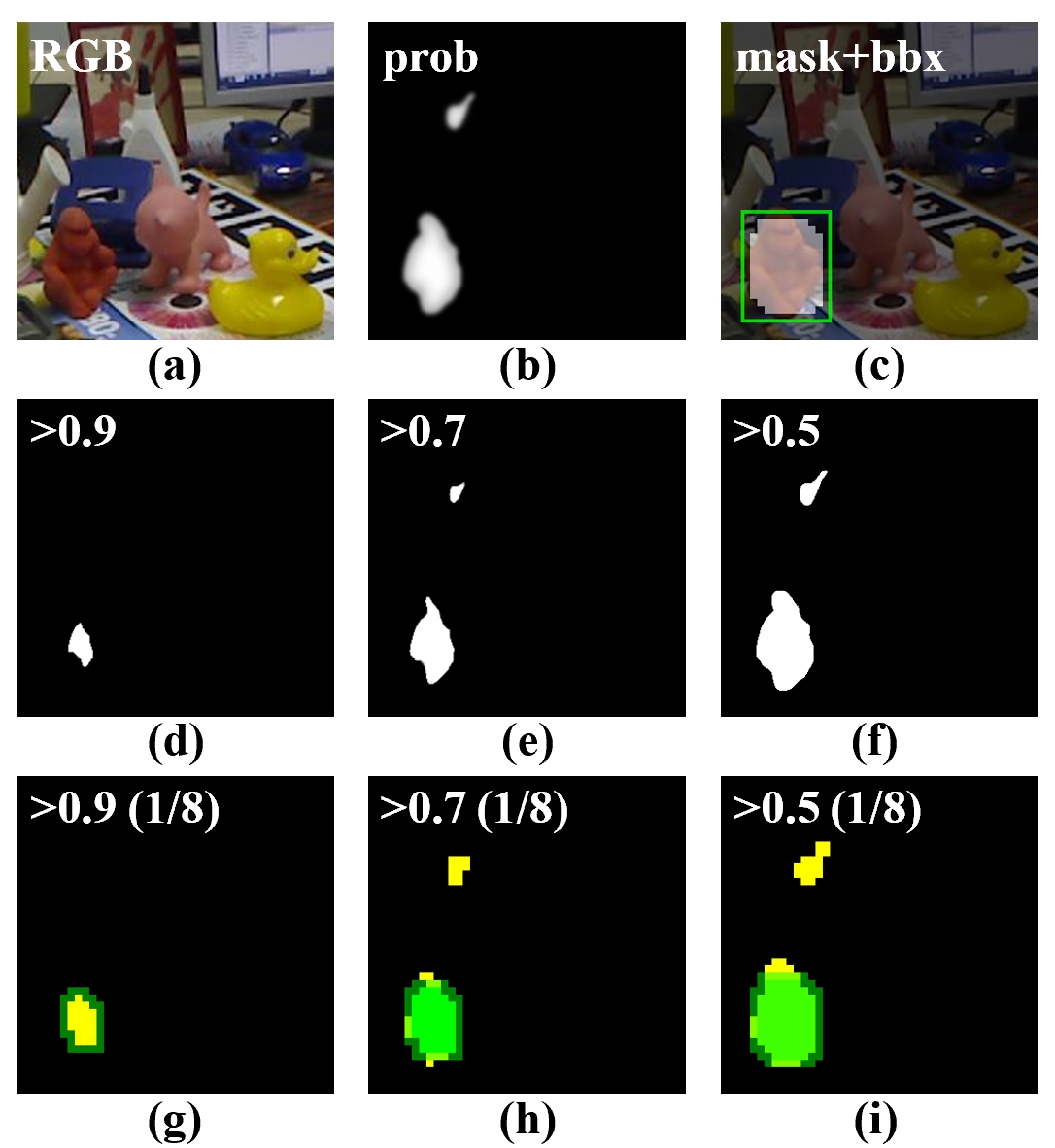}
      \caption{Visualization of the mask selection process. (a) is the input scene RGB image, (b) is the segmentation result which goes through softmax and comes out as probability map, (c) is the final masked area with bounding box. (d)(e)(f) are the masks derived by applying thresholds of 0.9, 0.7, 0.5 to the probability map. (g)(h)(i) are the max pooling map (by a factor of 8), yellow area is the original max pooling area, light green area is the overlapped area, dark green area is the expanded selected area.}
      \label{mask_selection}
    \end{figure}

The direct output of this segmentation part goes through a softmax process to produce probability map with a range of 0 to 1. Then, this map is thresholded at 0.5, 0.7 and 0.9 levels to obtain binary masks. Lower thresholds tend to retain false positive regions with potentially confusing features in addition to the true positive target areas. In contrast, higher thresholds may exclude portions of the true positive region where the model is less confident in its predictions. To leverage the complementary strengths of both cases, we employ a sequential region growing approach between the thresholds (see Fig. \ref{mask_selection})

Specifically, the thresholding is combined with a 8x max pooling for computational efficiency. Starting from the 0.9 mask, we incrementally expand the mask region by one pixel (equivalent to 8$\times$8 pixel area in the original mask) to cover more potential candidate regions. This expanded area is then overlapped with the 0.7 mask and expanded again, with the process repeating for the 0.9 mask. As evident in Fig. \ref{mask_selection}, this targeted expansion allows inclusion of dispersed true positive region while avoiding overlaps with isolated false positive regions that appear at lower thresholds. 

The bounding box that encloses the coarse mask with another one pixel expansion, is computed at the lower resolution and scaled back to the original resolution. It is used to crop out the corresponding region in the scene image. This crop, along with the associated contours and mask, are resized to a standardized resolution (currently set to \(128\times128\)), with padding to preserve the original aspect ratio. This processed patch serves as the input to the color-code estimator module.

The mask representations in the color-code estimator differ between the single and multi-object cases. In the single object estimation framework, only the mask of the target class is used for color-code estimation. However, when estimating the poses of multiple objects simultaneously, the coarse mask for one object may not be precise enough to exclude other objects, and providing a mask without indicating the object type could lead to confusion for the color-code estimator. To address this issue, a stack of mask layers for all the objects is fed into the color-code estimator, which shows the probabilities for each class in the cropped area and encodes the specific class for which the color-code is being predicted.

\subsection{Sparse Color Code Estimation} 
The color-code estimator adopts a UNet-style architecture as the segmentation module, with base filter numbers tuned for the task complexity. For single object mapping, the base filter number is set to 32, while multi-object scenarios increase this capacity to 48 filters to handle the greater complexity.

\begin{figure}[t]
        \centering
        \includegraphics[scale=0.14]{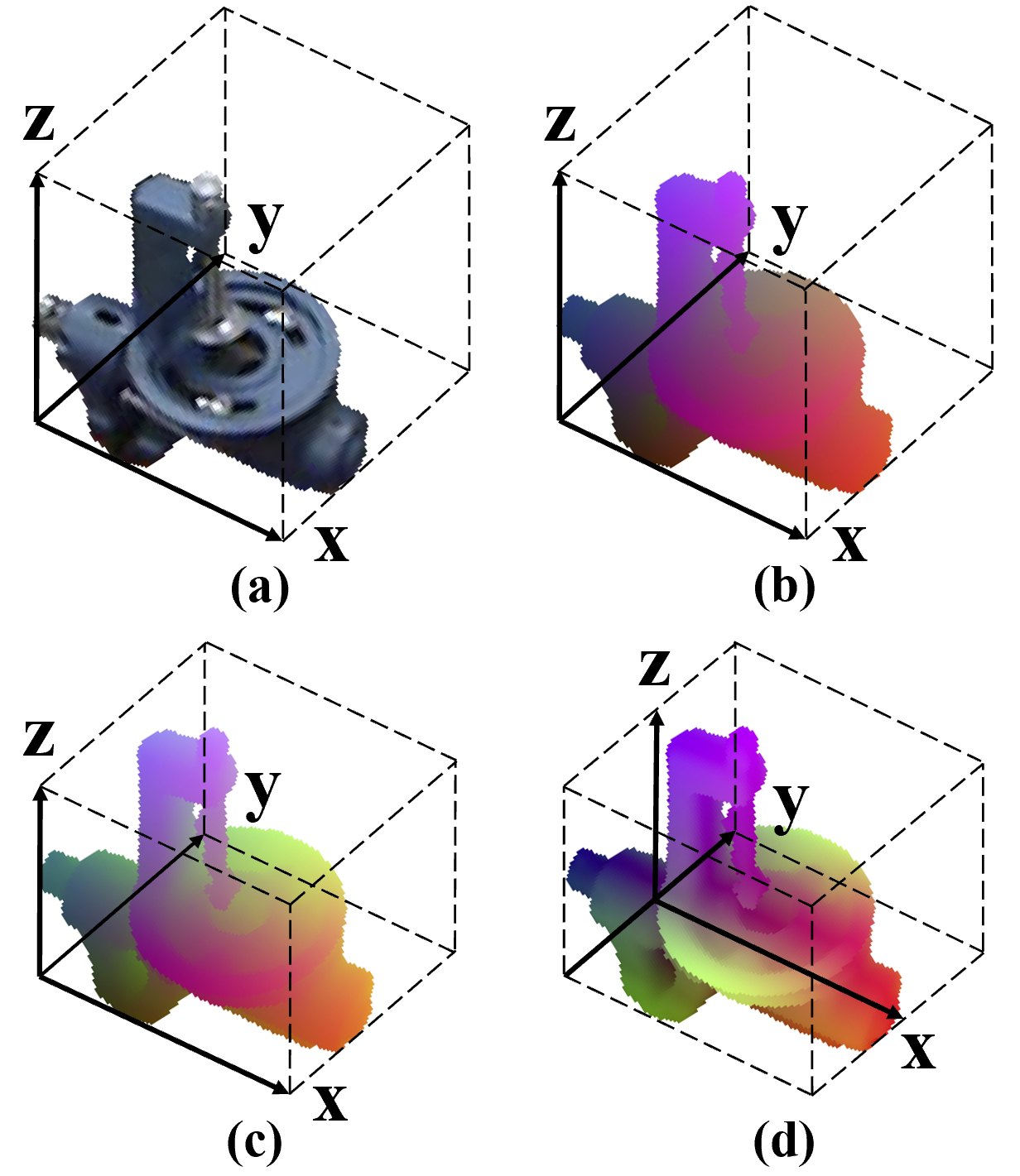}
        \caption{Different color-code visualization. The standard color-code (b)  normalizes the object to fit within a $1 \times 1 \times 1$ cube, with its maximum dimension spanning the full color channel range. The anisotropic color-code (c) allows each dimension to occupy the full range of its corresponding color channel. For objects with reflective symmetry, the symmetric anisotropic color-code (d) enables each symmetric part to span the full range of the respective color channel.}
        \label{colorcode_versions}
    \end{figure}

Standard color-codes are represented as normalized RGB cubes with equal XYZ extents (Fig. \ref{brief_illustration} left). However, when object geometry is imbalanced along the axes (e.g., Fig. \ref{colorcode_versions} showing a larger height range compared to width), this uniform mapping visually compresses the color-code. The resulting limited spatial color variations can make distinguishing neighboring object regions challenging. To address this issue, we propose an anisotropic color-code variant that redistributes the 0-255 RGB gamut to span the full XYZ ranges of the object. As evident in Fig. \ref{colorcode_versions}, this exaggerated mapping intensifies inter-region color differences, assisting networks in resolving correspondences between distinct surface features that may otherwise be confused in the default compressed encoding.

The sparse contour mask produced by the Sobel filter simplifies the color-code estimation task by explicitly highlighting the locations of the significant surface features. As noted previously, networks can struggle to resolve mappings in regions lacking differentiating cues from the surrounding areas. To direct model capacity towards these crucial regions, training applies higher loss weights to contours versus the rest, thereby emphasizing the errors in salient regions while retaining the overall color-code coherence across the object. Specifically, the color-code loss integrates L1 norms with regional weighting, formally defined as:
\begin{equation}
\label{eq:loss_cc}
Loss_{cc} = ||(I_{cc} - \hat{I}_{cc})||_1 +\lambda_{cntr} \cdot ||I_{cntr} \cdot (I_{cc} - \hat{I}_{cc}) ||_1 .
\end{equation}
Here, $I_{cc}$ is the ground-truth color-code, $\hat{I}_{cc}$ is the estimated color-code, $I_{cntr}$ is the contour mask, and $\lambda_{cntr}$ is the weight factor for the contour area that is set to 5 in our case.

\subsection{Symmetry Representation} 

In addition to predicting the color-code map, the estimator module is tasked with generating a symmetry mask for objects with reflective symmetry. If the symmetric object regions are encoded with entirely different color labels by default, this risks confusing the model. Consider horizontal flipping as a commonly used data augmentation technique in training neural networks as it improves model robustness by enforcing consistency between the original and flipped versions of the same input pattern.

To utilize symmetry as a consistent prior during training while avoiding feature confusion, a natural idea is mapping the color-code in a symmetric form. However, for subsequent pose estimation via PnP, symmetric regions need further separation into distinct 3D coordinates. Here, we propose a novel symmetry mask prediction along with a tailored loss capturing the separation of these symmetric parts. The loss for symmetry mask is defined as:
\begin{equation}
\label{eq:loss_symm}
% Loss_{symm} = \sum \left| I_{symm} \right| -\left| \sum(I_{symm} \cdot \hat{I}_{symm}) \right|
Loss_{symm} = \sum \left| I_{symm} \right| - \left| \sum(I_{symm} \cdot \hat{I}_{symm}) \right|
\end{equation}
where $I_{symm}$ is the ground-truth symmetry mask, and $\hat{I}_{symm}$ is the estimated symmetry mask.

% \textbf{\color{the || appears to be in different size?}}

\begin{figure}[b]
        \centering
         \includegraphics[scale=0.105]{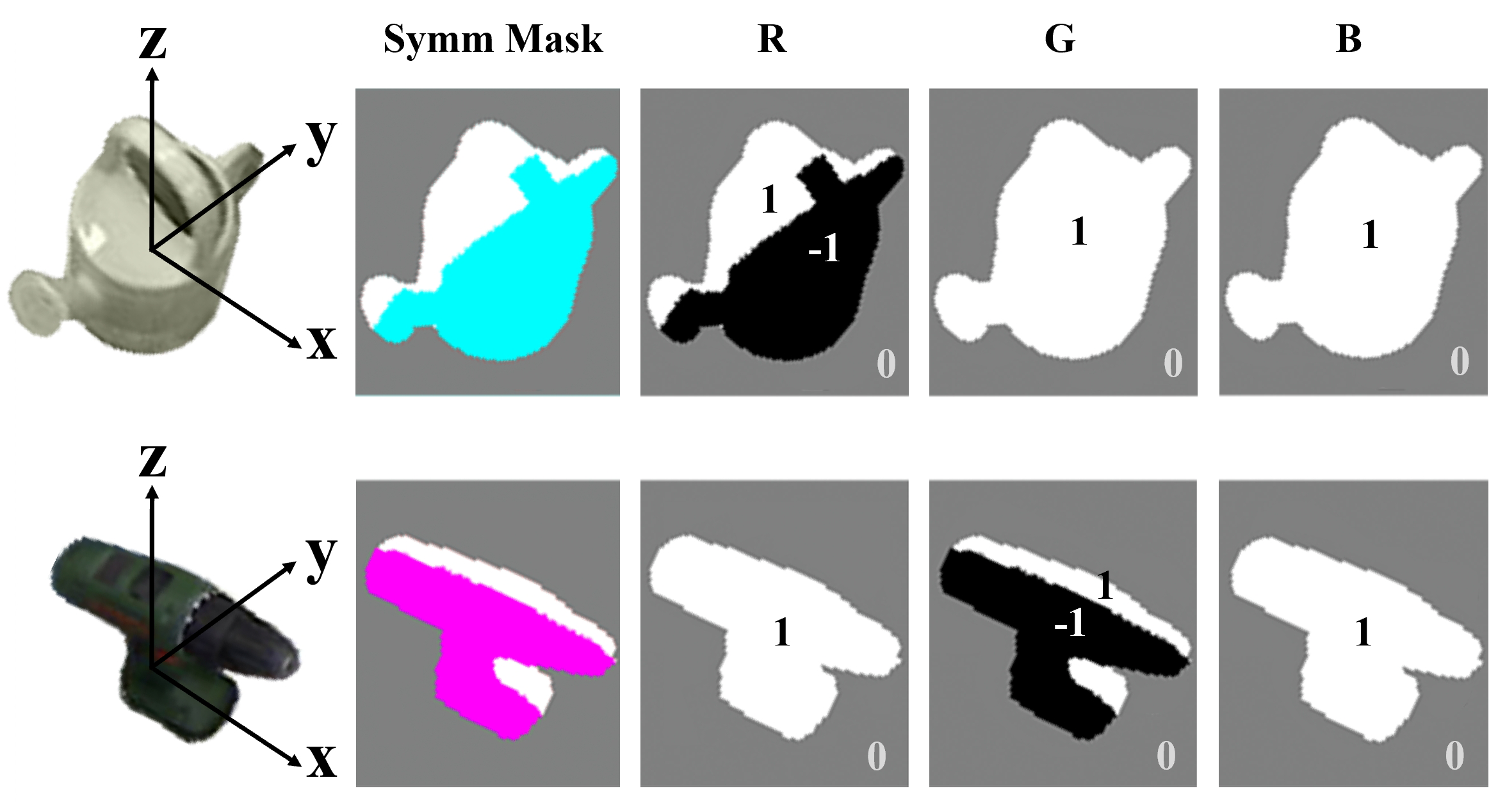}
        \caption{Symmetry mask visualization. The background is set to 0. If symmetry exists in a particular dimension (corresponding to a specific color channel), one of the object's symmetric part areas is assigned a value of 1, while the other is assigned -1. For the remaining color channels, the entire object area is assigned a value of 1.}
        \label{symmetry_mask_illustration}
    \end{figure}

Specifically, the symmetry mask labels the two reflective object areas with 1 and -1. While computing the loss, the pixelwise multiplication of the predicted and ground-truth masks are added together before taking the absolute value (Fig. \ref{symmetry_mask_illustration}, Eq. (\ref{eq:loss_symm})). This ensures that the loss of estimating the ground truth-aligned symmetry ($I_{symm}=1$ matches with $\hat{I}_{symm}=1$ and $I_{symm}=1$ matches with $\hat{I}_{symm}=-1$) is identical to the opposite case ($I_{symm}=1$ matches with $\hat{I}_{symm}=-1$ and $I_{symm}=-1$ matches with $\hat{I}_{symm}=1$). Therefore, the loss promotes local estimate consistency without explicitly encoding symmetry constraints, and guarantees a smooth gradient descent even if the model does not follow a consistent prediction of which area is labeled 1 or -1.
As the symmetry mask and the color-code estimate contain the same object region data, having the color-code estimator simultaneously predict both the outputs enables information sharing. 
%allows utilizing the shared information between them.

\subsection{Pose Estimation} 
Many pose estimation frameworks rely on locating projected 3D object points in the 2D image, making PnP a necessary step to determine pose. However, standard PnP uses Random sample consensus (RANSAC) for robustness against outliers. With large point sets derived from dense color-code prediction, this incurs heavy computational cost. Unlike the highly parallelizable neural network computation on GPUs, PnP can easily become the runtime bottleneck. Its non-linear complexity growth with added points contrasts the efficient deep network inference.

    \begin{figure}[t]
      \centering
      \includegraphics[scale=0.1]{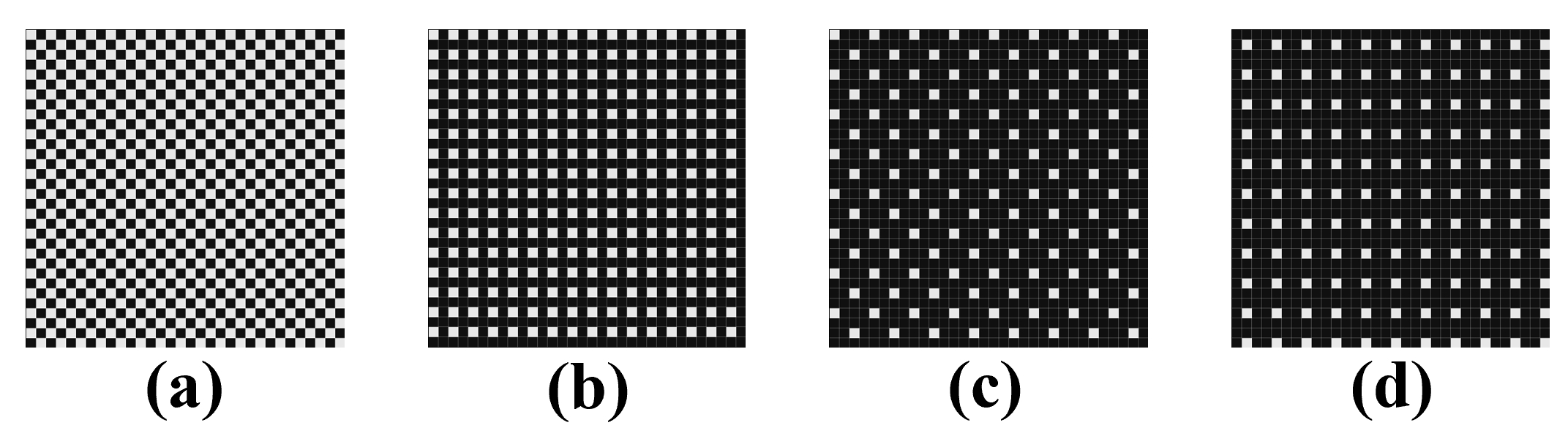}
      \caption{Masks generated using various sampling rates: (a) $1/2$, (b) $1/4$, (c) $1/8$, (d) $1/9$.}
      \label{sample_mask}
    \end{figure}

The Sobel contours extract pixels on distinct surface features. Using only these informative points for PnP maintains accuracy while reducing the computational burden of RANSAC iterations. If the initial candidate set is still too large, we can further prune it by evenly subsampling from the sparse contours. Based on a designated maximum point budget for PnP, different levels of sparsity can be applied (Fig. \ref{sample_mask}). This strategic point filtering balances pose fidelity and efficiency given performance targets.

For symmetric objects, when decoding the RGB color values back to 3D coordinates oriented at the default pose, the color-code pixels marked with -1 in the symmetry mask have their X/Y/Z values negated along the symmetrical axis. This mirrors asymmetric points across the plane of symmetry. Given 2D-3D point correspondences between image projections and default 3D coordinates, along with pre-calibrated camera intrinsic parameters, PnP estimation solves for the 6 degree-of-freedom rotation and translation relating the model points to the observed camera view.

\section{EXPERIMENTS}

In this section, we first describe our experimental configuration, implementation details, and evaluation methodology. The results are reported on two commonly used benchmark datasets, comparing to prior real-time capable techniques. Additionally, ablation studies quantify the contributions of the individual system components and the effectiveness of our approach is validated through a real-world experiment.
%on the LINEMOD dataset.

\subsection{Datasets} 
\textbf{LINEMOD \cite{hinterstoisser2013model}} consists of 13 sequences, each dedicated to one of the 13 objects, annotated with pose information and spanning approximately 1,200 images per object. This dataset is characterized by its cluttered settings and variable lighting conditions. Following established practices, roughly 15\%\ of the images are allocated for training purposes, with the remainder designated for testing.

\textbf{LINEMOD OCCLUSION \cite{brachmann2014learning}} is a subset of the LINEMOD dataset consisting of 1,214 heavily occluded images of 8 objects with pose annotations. It serves as an additional challenging test set for the LINEMOD objects, and provides extra ground-truth pose labels for modeled objects under significant inter-object occlusions. By evaluating pose estimation approaches on real test cases with heavy occlusions, it benchmarks the capability to infer poses despite limited observations and overlapping objects.

\begin{table*}[b!]
\centering
\caption{EXPERIMENT ON LINEMOD DATASET. *Indicates symmetric objects evaluated with ADD-S metric, \dag indicates symmetric objects evaluated with ADD-S' metric}
\label{tab:linemod}
\resizebox{\textwidth}{!}{
\begin{tabular}{l|l|l|l|l|l|l|l|l|l}
  \hline
& PVNet\cite{peng2019pvnet} & SingleShot\cite{tekin2018real} & PoseCNN\cite{xiang2017PoseCNN} & Pix2Pose\cite{park2019Pix2Pose} & CDPN\cite{li2019cdpn} & DPOD\cite{zakharov2019dpod} & HybridPose\cite{song2020hybridpose} & COPE\cite{thalhammer2023cope} & SCCN (Ours) \\
  \hline
ape & 43.62 & 21.62 & 27.8 & 58.1 & 64.38 & 53.28 & 63.1 & - & 61.8 \\
benchwise\dag & 99.9 & 81.8 & 68.9 & 91 & 97.77 & 95.34 & 99.9 & - & 84.7 \\
cam & 86.86 & 36.57 & 47.5 & 60.9 & 91.67 & 90.36 & 90.4 & - & 66.7 \\
can\dag & 95.47 & 68.8 & 71.4 & 84.4 & 95.87 & 94.1 & 98.5 & - & 81.3 \\
cat & 79.34 & 41.82 & 56.7 & 65 & 83.83 & 60.38 & 89.4 & - & 72.5 \\
driller\dag & 96.43 & 63.51 & 65.4 & 76.3 & 96.23 & 97.72 & 98.5 & - & 68.7 \\
duck\dag & 52.58 & 27.23 & 42.8 & 43.8 & 66.76 & 66.01 & 65 & - & 46.1 \\
eggbox*\dag & 99.15 & 69.58 & 98.3 & 96.8 & 99.72 & 99.72 & 100 & - & 93.5 \\
glue*\dag & 95.66 & 80.02 & 95.6 & 79.4 & 99.61 & 93.83 & 98.8 & - & 80.4 \\
holepuncher\dag & 81.92 & 42.63 & 50.9 & 74.8 & 85.82 & 65.83 & 89.7 & - & 80.2 \\
iron\dag & 98.88 & 74.97 & 65.6 & 83.4 & 97.85 & 99.8 & 100 & - & 82.9 \\
lamp\dag & 99.33 & 71.11 & 70.3 & 82 & 97.89 & 88.11 & 99.5 & - & 84.3 \\
phone & 92.41 & 47.74 & 54.6 & 45 & 90.75 & 74.24 & 94.9 & - & 62.6 \\
  \hline
mean & 86.27 & 55.95 & 62.75 & 72.4 & 91.36 & 82.98 & 91.3 & 73.8 & 74.3 \\
  \hline
\end{tabular}
}
\end{table*}

\subsection{Evaluation Metrics}
The ADD metric for 6D object pose estimation calculates the average distance between corresponding points on the 3D model of an object, transformed by the ground truth and predicted poses. The estimation is considered correct if the average distance is less than 10\%\ of the object's diameter.
\begin{equation}
\label{eq:add}
ADD = \frac{1}{m} \sum_{x \in \mathcal{M}} \| (Rx + t) - (\hat{R}x + \hat{t})\|
\end{equation}
where \( m \) is the number of points in the model \( \mathcal{M} \), \( R \) and \( t \) are the ground truth rotation and translation, respectively, and \( \hat{R} \) and \( \hat{t} \) are the predicted rotation and translation.

Traditionally, ADD-S score \cite{kehl2017ssd} has been used for a few of the symmetric objects, namely, eggbox and glue, in the LINEMOD dataset. Here we expand the list of symmetric objects to benchwise, can, driller, duck, eggbox, glue, holepuncher, iron, and lamp. For these objects with reflective symmetry, we propose a ADD-S' score for pose evaluation. This ADD-S' score is calculated by taking the minimum value between the scores obtained from the direct pose and the reflected pose as:
\begin{equation}
\label{eq:add-s}
\begin{split}
ADD\text{-}S' = \frac{1}{m} \sum_{x \in \mathcal{M}} \min \Big(&\| (Rx + t) - (\hat{R}R_{refl}x + \hat{t})\|,\\
&\| (Rx + t) - (\hat{R}x + \hat{t})\|\Big) 
\end{split}
\end{equation}
where $R_{refl}$ is the rotation matrix to account for reflective symmetry.

\subsection{Implementation Details} 
We implement our model in PyTorch and train it on an Intel i9-7900X CPU and NVIDIA RTX3090 GPU. Runtime measurements are performed on an NVIDIA Jetson AGX Xavier. The segmentation section and color-code estimator are trained with a batch size of 16 for 150 epochs, using the Adam optimizer. The segmentation section has an initial learning rate of 0.05, while the color-code estimator starts at 0.001. Both the sections employ an Adaptive Learning Rate Scheduler \cite{chen2022bootstrap} to dynamically adjust the learning rate during training. Specifically, if the average loss of the current epoch is more than 98\% of the loss from the previous epoch, the learning rate for the next epoch is multiplied by 0.98.

We synthesize an additional 3,000 images per LINEMOD object using the PVNet \cite{peng2019pvnet} rendering approach to expand the training data. The Neural 3D Mesh Renderer \cite{kato2018neural} is used to generate the corresponding ground-truth color-coded images. Data augmentation is performed with two TorchVision Transform groups - photometric and geometric. The photometric transforms, applied only to the RGB scenes, includes RandomPosterize, RandomAdjustSharpness, ColorJitter (modifying brightness, contrast, saturation and hue) with a multiplier of 5 for the 15\% LINEMOD split and 3 for the rendered images. The geometric transforms, applied to RGB scenes and the corresponding color-code and symmetry mask images, consists of RandomCrop, RandomPerspective and Resize. 
%This dual data augmentation strategy artificially expands the diversity of scenes and object poses to improve model generalization. 
For multi-object pose estimation under occlusions, we apply a cut-and-paste strategy to generate 10,000 augmented images with half from the LINEMOD dataset and half from the images obtained from PVNet rendering. This expands the diversity of inter-object occlusions for robust training. Additionally, during color-code estimator training, we randomly apply geometric transforms to the object masks and then overlay these masks onto the target objects to simulate occlusions. By intentionally masking portions of the target objects in different locations and shapes, we enhance the model's capability to infer poses despite partial observations due to scene obstructions.

\subsection{Evaluation Results}

\begin{table*}[h!]
\centering
\caption{EXPERIMENT ON LINEMOD OCCLUSION DATASET}
\label{tab:linemod-occ}
\resizebox{\textwidth}{!}{
\begin{tabular}{l|l|l|l|l|l|l|l|l} % 9 columns including the first column for item names
\hline
& PVNet\cite{peng2019pvnet} & PoseCNN\cite{xiang2017PoseCNN} & Pix2Pose\cite{park2019Pix2Pose} & HybridPose\cite{song2020hybridpose} & Seg-Driven6D\cite{hu2019segmentation} & COPE\cite{thalhammer2023cope} & GDR-Net\cite{wang2021gdr} & SCCN (Ours) \\
\hline
ape & 15.81 & 9.6 & 22 & 20.9 & 12.1 & - & 41.3 & 20.3 \\
can & 63.3 & 45.2 & 44.7 & 75.3 & 39.9 & - & 71.1 & 45.2 \\
cat & 16.68 & 0.93 & 22.7 & 24.9 & 8.2 & - & 18.2 & 21.7 \\
driller & 65.65 & 41.4 & 44.7 & 70.2 & 45.2 & - & 54.6 & 47.8 \\
duck & 25.24 & 19.6 & 15 & 27.9 & 17.2 & - & 41.7 & 17.4 \\
eggbox & 50.17 & 22 & 25.2 & 52.4 & 22.1 & - & 40.2 & 30.3 \\
glue & 49.62 & 38.5 & 32.4 & 53.8 & 35.8 & - & 59.5 & 34.5 \\
holepuncher & 39.67 & 22.1 & 49.5 & 54.2 & 36 & - & 52.6 & 47.7 \\
\hline
mean & 40.77 & 24.9 & 32 & 47.5 & 27 & 35 & 47.4 & 33.1 \\
\hline
\end{tabular}
}
\end{table*}

\begin{table*}[h!]
\centering
\caption{SPEED PERFORMANCE ON NVIDIA JETSON AGX XAVIER}
\label{tab:speed}
\resizebox{\textwidth}{!}{
\begin{tabular}{l|l|l|l|l|l|l|l|l|l} 
\hline
& PVNet\cite{peng2019pvnet}& SingleShot\cite{tekin2018real} & DPOD\cite{zakharov2019dpod} & Pix2Pose\cite{park2019Pix2Pose} & HybridPose\cite{song2020hybridpose} & Seg-Driven6D\cite{hu2019segmentation} & COPE\cite{thalhammer2023cope} & GDR-Net\cite{wang2021gdr} & SCCN (Ours) \\
\hline
LINEMOD & 3.7     & 8.3  & 4.8   & 1.5     & 2.2     & 4.3 & 1.9 & 4.4 & \textbf{19} \\
LINEMOD-OCC & 3.7/obj  & 5.2  & 3.1   & 1.5/obj & 2.2/obj & 2.4 & 1.7 & 2.9 & \textbf{6.5} \\
\hline
\end{tabular}
}
\end{table*}

\begin{table}[h!]
\centering
\caption{ABLATION STUDY}
\label{tab:ablation}
\resizebox{\columnwidth}{!}{
\begin{tabular}{l|l|l|l|l}
\hline
& standard & aniso & aniso + symm & aniso + symm (sparse) \\
\hline
benchwise & 77.2 & 83.4 & 85.3 & 84.7 \\
can & 72.3 & 79.5 & 82.6 & 81.3 \\
driller & 61.9 & 67.3 & 69.1 & 68.7 \\
duck & 42.4 & 45.2 & 46.6 & 46.1 \\
eggbox & 84.4 & 91.5 & 94.9 & 93.5 \\
glue & 74.5 & 80.0 & 81.7 & 80.4 \\
holepuncher & 73.6 & 79.1 & 81.2 & 80.2 \\
iron & 75.8 & 80.7 & 83.4 & 82.9 \\
lamp & 77.5 & 82.3 & 85.7 & 84.3\\
  \hline
mean        & 71.1 & 76.6 & 78.9 & 78.0 \\
  \hline
\end{tabular}
}
\end{table}

We evaluate our method's performance against other methods that claim to obtain real-time performance and also have open-source code which makes evaluation on NVIDIA Jetson AGX Xavier feasible. 
As shown in Table \ref{tab:linemod}, our method achieves an average ADD score of 74.2 across all the LINEMOD objects for pose estimation. While this does not surpass the highest recorded ADD scores, our result remains comparable to other techniques given the efficiency of our approach as our model strikes a balance between accuracy and simplicity. The decent level of accuracy attained despite efficiency constraints demonstrates the effectiveness of the proposed colored-code design for extracting pose-relevant features. 
Table \ref{tab:linemod-occ} shows that our approach achieves an average ADD score of 33 on the LINEMOD OCCLUSION dataset. Even though this score is not the highest, it remains reasonably competitive compared to other techniques. Given the efficiency of our model, the accuracy attained demonstrates effectiveness at pose estimation even in heavily occluded scenes with multiple objects.

To provide an objective and meaningful assessment of practical (real-world) runtime efficiency, we benchmark all the pose estimation methods on an NVIDIA Jetson Xavier edge platform. The average inference speeds are shown in Table \ref{tab:speed}. SCCN achieves 19 FPS for single object pose estimation and 6.5 FPS for simultaneous predictions of 8 objects in a single frame, both of which are substantially more than the state-of-the-art. The performance is further accelerated by transforming the model to the TensorRT format with single and multiple object estimations reaching 26 FPS and 9 FPS, respectively. Considering that some of the top-scoring techniques use end-to-end architectures, the real-time capability and decent accuracy attained by our lightweight 3-stage pipeline is significant.

We investigate the impact of various enhancements through an ablation study, which includes the standard color-code, anisotropic color-code, anisotropic color-code with symmetry representation, and color-code with sparse input. Only 9 of the 13 objects from the LINEMOD dataset are selected for the comparison as they are considered to have reflective symmetry. Table \ref{tab:ablation} demonstrates that leveraging sparsity does not significantly impact estimation accuracy, while both the anisotropic and symmetric color-code representations provide a noticeable improvement over the standard color-code version.

    \begin{figure}[b]
        \centering
         \includegraphics[scale=0.14]{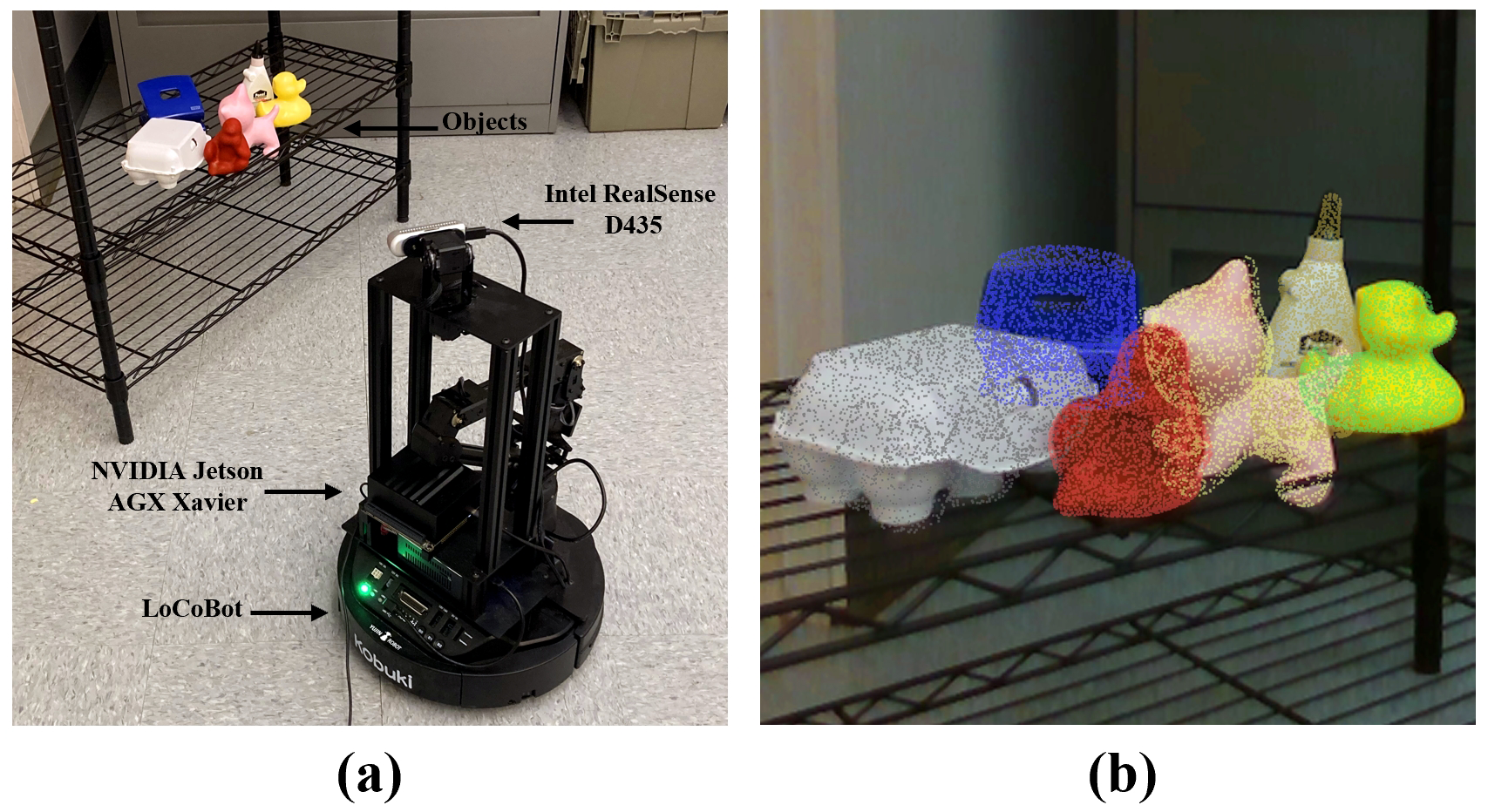}
        \caption{Qualitative results on experiment. (a) Real-world experimental setup with LoCoBot platform. (b) Qualitative result visualized by projecting the point clouds (sampled from the object mesh file) onto the image using the predicted object pose and camera intrinsics parameters.}
        \label{experiment}
    \end{figure}

\subsection{Proof-of-concept Demonstration}
We deploy the TensorRT-optimized pipeline on a LoCoBot platform equipped with an Intel RealSense D435 camera, controlled via the PyRobot interface. The SCNN network processes the camera images on an on-board NVIDIA Jetson AGX Xavier device, which features a 512-core Volta GPU with Tensor Cores and an 8-core ARM v8.2 64-bit CPU. Fig. \ref{experiment} shows the experimental setup and depicts a representative qualitative result, wherein the projected object point clouds, generated based on their 6D pose estimates, align well with the corresponding object images.

\section{CONCLUSIONS}

In summary, our 6D object pose estimation framework, SCCN, achieves a desirable balance of precision and speed for responsive (low latency) real-time systems, and demonstrates promising performance on an edge device under real-world constraints. Specifically, sparse color-code pixel inputs yield an efficient compute architecture to retain estimation accuracy while maximizing inference speed. The results also validate the usefulness of the proposed anisotropic and symmetric color-code representations. 
In the future, we plan to extend SCCN's capability to estimate the poses for multiple instances of the same object class simultaneously. Additionally, we aim to enhance the model's generalization ability, enabling it to accurately predict the poses of novel objects and adapt to test domains that differ substantially from the training domains. We also intend to integrate SCCN with our robust object recognition \cite{samani2021visual,samani2024persistent} and probabilistic mapping frameworks \cite{wong2023human} for effective mobile manipulation in cluttered spaces.

\section*{ACKNOWLEDGMENT}
We acknowledge the use of ChatGPT in light-editing of our text. 

\bibliographystyle{IEEEtran}
\bibliography{refs}

\end{document}